\documentclass[11pt]{article}

\usepackage[final]{acl}

\usepackage{times}
\usepackage{latexsym}

\usepackage[T1]{fontenc}

\usepackage[utf8]{inputenc}

\usepackage{microtype}

\usepackage{inconsolata}

\usepackage{graphicx}
\usepackage[utf8]{inputenc}

\usepackage{amsmath, amsfonts, amsthm, amssymb, graphicx}
\usepackage{mathtools,verbatim}
\usepackage{enumitem}
\usepackage{natbib}
\usepackage{algorithm, algpseudocode}
\usepackage{hyperref}
\usepackage{fancyhdr}
\usepackage[capitalize,noabbrev]{cleveref}
\AddToHook{cmd/appendix/before}{%
    \crefalias{section}{appendix}%
    \crefalias{subsection}{appendix}
}
\usepackage{tikz}
\usepackage{multirow}
\usepackage{booktabs}
\usepackage{hhline}
\usepackage{array}
\usepackage{lipsum}
\usepackage{multirow}
\usepackage[table]{xcolor}
\usepackage{subcaption}

\numberwithin{equation}{section}

\renewcommand{\hbar}{\tau}

{
 \theoremstyle{plain}
      
}
\newtheorem{nono-theorem}{Theorem}[]
\theoremstyle{plain}

\theoremstyle{definition}

\title{Shared Low-rank Basis Factorization for \\
Data-free Mixture-of-Experts Compression}

\author{
  \textbf{Tianxiao Cao\textsuperscript{1}\thanks{
  Work done primarily while at RIKEN AIP.
  }},
  \textbf{Jiahe Shao\textsuperscript{3}},
  \textbf{Yuning Qiu\textsuperscript{2}},
  \\
  \textbf{Kyohei Atarashi\textsuperscript{1}},
  \textbf{Hisashi Kashima\textsuperscript{1}},
  \textbf{Qibin Zhao\textsuperscript{2}}
  \\
  \textsuperscript{1}Kyoto University,
  \textsuperscript{2}RIKEN AIP,
  \textsuperscript{3}The University of Tokyo
  \\
  \texttt{cao.tianxiao.65w@st.kyoto-u.ac.jp}
}

\begin{document}
\maketitle
\begin{abstract}
Mixture-of-Experts (MoE) large language models decouple capacity from compute through sparse routing, but their large parameter count creates storage and serving challenges. We analyze three MoE compression families: expert pruning, expert merging, and weight reconstruction, and derive structural error bounds showing that pruning and merging can incur non-vanishing errors tied to routing and expert heterogeneity. In contrast, weight reconstruction avoids these structural costs by preserving expert structure and routing. Motivated by the analysis, we propose Shared Low-rank Basis Factorization (SLBF), a data-free weight reconstruction method that uses rank-$k$ bases shared among experts, enabling richer cross-expert sharing, faster convergence, and lower reconstruction error. A post-hoc gauge fixing removes redundant parameters at no representational cost. Across five MoE architectures spanning 16B to 122B parameters, SLBF consistently outperforms methods from all three compression families.
\end{abstract}

\section{Introduction}

The Mixture-of-Experts (MoE) architecture~\citep{jacobs1991adaptive,shazeer2017} has become the dominant paradigm for scaling large language models, adopted by a rapidly growing family of production LLMs~\citep{jiang2024mixtral,liu2024deepseek,qwen3technicalreport,gemma4}. By decoupling total capacity from compute through sparse routing, MoE enables models with hundreds of billions of parameters while maintaining efficient inference.

Sparse activation solves the compute problem but not the memory problem. Every expert must reside in accelerator memory regardless of how rarely it is activated: Qwen3-235B-A22B deploys 128 expert FFNs per layer yet routes each token to only 8, leaving over 93\% of expert parameters idle per forward pass but occupying accelerator memory. At larger scales, the situation is acute: deploying DeepSeek-V3 (671B) requires distributing expert shards across dozens of GPUs, and the storage and transfer cost of expert weights dominates both the serving budget and the model distribution pipeline. To reduce their memory footprint while preserving the routing structure that enables sparse activation, compressing the expert parameters of MoE-based LLMs is therefore a practical and urgent problem.

Existing approaches to MoE compression intervene at one of three structural levels: expert pruning~\citep{lasby2026reap}; expert merging~\citep{chen2025hcsmoe}; and weight reconstruction~\citep{liu2025molae,chen2026mobe}. We derive structural error bounds for pruning and merging, and show that both can incur non-vanishing errors tied to routing and expert heterogeneity in modern MoE architectures where load balancing encourages broad expert utilization and expert specialization induces functional diversity. By preserving both expert structure and routing, weight reconstruction avoids these intervention-induced structural costs, while its remaining error arises from imperfect parameter reconstruction.

Within the weight reconstruction family, the state-of-the-art MoBE formulation~\cite{chen2026mobe} shows significant strengths by parameterizing expert weights as non-linear mixtures of shared full-rank bases. However, scaling these full-rank bases rapidly consumes the parameter budget, creating a bottleneck that limits the model's representational capacity. 
To resolve this limitation, we propose Shared Low-rank Basis Factorization (SLBF), which replaces each full-rank basis with a rank-$k$ factorization $\{\mathbf U_{j} \mathbf V_{j}\}_{j=1}^{m}$. The lower per-basis cost allows more bases under the same compression budget. This shift from a few full-rank bases to many low-rank bases enables faster convergence, a richer mixing structure where experts compose across multiple bases rather than concentrating on a single dominant one, and ultimately lower weight reconstruction error. A post-hoc gauge fixing eliminates $mk^{2}$ redundant parameters at no representational cost. SLBF is entirely data-free and requires no fine-tuning after compression. In summary, our main contributions are:

\begin{itemize}
    \item We derive a unified structural cost analysis of three MoE compression families, with new error bounds for expert pruning and weight reconstruction, showing that weight reconstruction avoids the intervention-induced structural costs of pruning and merging.
    \item We propose the Shared Low-rank Basis Factorization with bilinear gauge fixing that enables richer cross-expert basis sharing under fixed compression budgets.
    \item We evaluate SLBF on five MoE architectures spanning 16B to 122B parameters against competing methods from all three families, demonstrating consistent improvements.
\end{itemize}
\section{Related Work}\label{sec:related-work}

\paragraph{MoE architecture.} 
Originally proposed by~\citet{jacobs1991adaptive} and revived for deep learning by~\citet{shazeer2017}, the sparse MoE design enables scaling of parameters without proportional increase in compute~\citep{lepikhin2021gshard,fedus2022switch}. 
Modern MoE-based LLMs adopt MoE FFNs~\citep{jiang2024mixtral,liu2024deepseek,qwen3technicalreport,gemma4} with refinements including load balancing~\citep{wang2024auxiliary}, shared experts~\citep{dai2024deepseekmoe}, and fine-grained granularity with hundreds of experts per layer~\citep{qwen3technicalreport}. These developments have made MoE a leading architecture for scaling LLM capacity.

\paragraph{Expert merging.}
Expert merging groups experts and combines the weights within each group into shared representatives, reducing the number of distinct expert parameter sets. \citet{li2024mcsmoe} groups experts by router-logit similarity and merges via frequency-weighted averaging;~\citet{chen2025hcsmoe} improves clustering quality through hierarchical agglomerative clustering on expert output vectors;~\citet{jha2026ream} improves grouping quality by using activation-based saliency scores. 

\paragraph{Expert pruning.} 
Expert pruning reduces MoE model size by removing entire experts based on a saliency criterion. Early work required task-specific fine-tuning~\citep{chen2022task}, while subsequent methods operate in retraining-free settings using criteria such as router frequency, reconstruction loss minimization~\cite{lu2024naee}, gradient-free search~\cite{liu2024efficient}, continuous relaxation of the pruning decision~\citep{bai2025diep}, and expert activation norms~\citep{jaiswal2025ean}. REAP \citep{lasby2026reap} combines router values with activation norms into the pruning criterion.

\paragraph{Weight reconstruction.} 
Weight reconstruction compresses expert weight matrices while preserving all experts and routing. MoLAE~\citep{liu2025molae} applies SVD to represent each expert through a shared latent matrix; D$^2$-MoE~\citep{gu2025delta} decomposes weights into shared and delta components; MoBE~\citep{chen2026mobe} re-parameterizes each expert's right factor as a non-linear mixture of shared basis matrices.

\paragraph{Other compression techniques.}
Beyond the above three families, quantization~\citep{duanmu2025mxmoe,chen2025moequant} and weight pruning~\citep{xie2024moepruner} have also been adapted to MoE compression. Complementary post-pruning techniques, such as converting pruned experts into bias vectors~\citep{zong2026not} or replacing them with sub-networks~\citep{zhang2026mone}, can further compensate for compression loss. These approaches operate along a different axis from the interventions considered here and are outside the scope of our structural analysis.
\begin{figure*}[t]
    \centering
    \begin{subfigure}[t]{0.49\linewidth}
        \centering
        \includegraphics[width=\linewidth]{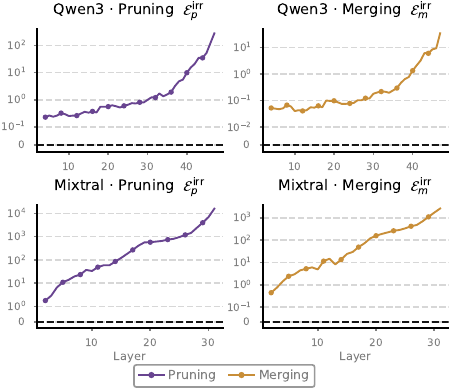}
        \caption{Pair-level structural costs.}
        \label{fig:irr-grid}
    \end{subfigure}
    \hfill
    \begin{subfigure}[t]{0.49\linewidth}
        \centering
        \includegraphics[width=\linewidth]{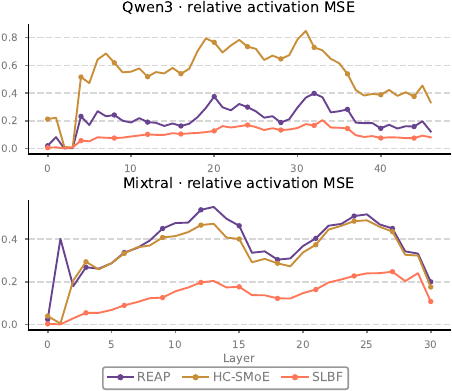}
        \caption{Realized per-layer compression error}
        \label{fig:hidden-layer}
    \end{subfigure}
    \caption{\textbf{(a)} \textbf{Pair-level structural costs} $\bar{\mathcal{E}}_p^{\mathrm{irr}}$ (pruning, left column) and $\bar{\mathcal{E}}_m^{\mathrm{irr}}$ (merging, right column) per MoE layer on the uncompressed model, averaged over expert pairs activated on the calibration set. \textbf{(b)} \textbf{Realized per-layer compression error} $\|\delta \mathbf h_\ell\|^2 / \|\mathbf h_\ell^{\mathrm{orig}}\|^2$ at 25\% MoE compression for REAP (pruning), HC-SMoE (merging), and SLBF (weight reconstruction). Top: Qwen3 ($n=128$, top-8). Bottom: Mixtral ($n=8$, top-2).}
    \label{fig:main}
\end{figure*}

\section{Structural Motivation}
\label{sec:motivation}

\paragraph{MoE setup.}

We focus on sparse Mixture-of-Experts (MoE) models. Each feed-forward network (FFN) in the Transformer is treated as one expert. Modern MoE-based LLMs~\cite{jiang2024mixtral,liu2024deepseek,qwen3technicalreport,gemma4} typically implement each expert as a SwiGLU-style FFN. 
A sparse MoE layer consists of a set of $n$ experts $\{f^i(\cdot)\}_{i=1}^n$ and a router.
Specifically, the $i$-th expert is formulated as:
\begin{equation}
    f^{i}(\mathbf{x}) = \left(\operatorname{SiLU}(\mathbf x \mathbf W^{i}_{\text{gate}}) \odot (\mathbf x \mathbf W^{i}_{\text{up}})\right) \mathbf W^{i}_{\text{down}},
\end{equation}
where $\mathbf{x}\in \mathbb R ^{d}$ is the input hidden state; $\mathbf W^{i}_{\text{gate}}$, $\mathbf W^{i}_{\text{up}}\in \mathbb R^{d \times d_{\text{ff}}}$, and $\mathbf W^{i}_{\text{down}} \in \mathbb R^{d_\text{ff}\times d}$ are the weight matrices. 
The router assigns hidden states to experts by applying a Top-$K$ selection over a linear projection of the hidden states, followed by a Softmax normalization. This router mechanism can be formulated as:
\begin{equation}
    \begin{aligned}
        \mathbf{g} &= \operatorname{Softmax}(\mathbf{x} \mathbf{W}_g), \\
        \mathcal{P}(\mathbf x) &= \operatorname{argTopK}(\mathbf{g}, K), \\
        \mathbf{h}(\mathbf x) &= \sum_{i \in \mathcal{P}} g_i f^i(\mathbf{x}),
    \end{aligned}
\end{equation}
where $\mathbf W_g\in \mathbb R^{d \times n}$ parameterizes the linear projection of the router.

\subsection{Three interventions, three structural costs}
\label{sec:cost-analysis}

\paragraph{Setup.} We analyze the structural cost of each family (pruning, merging, and weight reconstruction, as introduced in~\cref{sec:related-work}) in the simplified setting of compressing a pair of experts $(f^i, f^j)$ with respective routing $g_i(\mathbf x), g_j(\mathbf x)$, adopting the pairwise analysis framework of~\citet{lasby2026reap}.
Each compression family modifies the expert pair in a distinct way, giving rise to a different source of compression error. We summarize the three interventions below; detailed derivations are provided in~\cref{appendix:derivations}. We call the minimum residual under each specified pairwise intervention its irreducible structural cost, denoting those of pruning and merging by $\mathcal E^{\mathrm{irr}}_p$ and $\mathcal E^{\mathrm{irr}}_m$, respectively.

\paragraph{Merging (following~\citet{lasby2026reap}).}
The pair contribution $g_i(\mathbf x) f^i(\mathbf x) + g_j(\mathbf x) f^j(\mathbf x)$ admits the decomposition $(g_i + g_j)\bigl(r(\mathbf x) f^i + (1-r(\mathbf x)) f^j\bigr)$, where $r(\mathbf x) := g_i / (g_i + g_j) \in [0,1]$ is the input-dependent mixing ratio. Merging replaces $(f^i, f^j)$ with a single representative $\tilde f^{ij}$ and replaces their separate routing weights with the summed routing $(g_i + g_j)$. 
Even with the optimal static mixing coefficient $\alpha^\star$, a residual error remains because the input-dependent $r(\mathbf x)$ is replaced by a static mixture. Under a weak-correlation approximation of the three multiplicative factors, this structural cost is
\begin{equation}\label{eq:error-irr-m}
\mathcal E_m^{\mathrm{irr}} \approx \mathbb{E}_{\mathbf x}\bigl[(g_i + g_j)^2\bigr]\cdot \operatorname{Var}\bigl[r(\mathbf x)\bigr]\cdot \mathbb{E}_{\mathbf x}\|\Delta_{ij}\|^2,
\end{equation}
where $\Delta_{ij}(\mathbf x) := f^i(\mathbf x) - f^j(\mathbf x)$. 

\noindent\textit{Remark.} Under this approximation, the merging cost grows with the expert functional gap $\|\Delta_{ij}\|^2$, router variability $\operatorname{Var}[r]$, and pair routing mass $(g_i+g_j)^2$. It vanishes when routing is static on the pair ($\operatorname{Var}[r]=0$) or the experts are functionally identical ($\Delta_{ij}=0$).

\paragraph{Pruning.} We derive a novel bound for the pairwise pruning intervention. This intervention removes one expert from the pair $(f^i,f^j)$. Without loss of generality, 
suppose expert $j$ is removed and its contribution is compensated solely by the surviving expert $i$ through a modified router $g^\prime_i(\cdot)$
Even with the \textit{oracle} routing function $g^{\star}_i$ which minimizes the output loss after pruning expert $j$, we show that expert pruning exhibits an error of
\begin{equation}\label{eq:error-irr-p}
    \mathcal E^{\mathrm{irr}}_p = \mathbb E_\mathbf x\left[g^2_j \left(\|f_j\|^2-\frac{|\langle f_j, f_i\rangle|^2}{\|f_i\|^2}\right)\right].
\end{equation}
\textit{Remark.} The cost is positive when, on a non-trivial subset of inputs, expert-$j$ is active ($g_j > 0$) and its output is not parallel to the substitute's ($f_j \not\parallel f_i$).

\paragraph{Weight Reconstruction.} 
Unlike pruning and merging, weight reconstruction preserves both expert identities and routing, and instead compresses each expert weight tensor $\mathbf W_k$ through a reduced-parameter representation $\hat{\mathbf W}_k$. For simplified linear experts $f^k(\mathbf x) = \mathbf x \mathbf W^k $, the resulting output error is bounded directly by the router-weighted reconstruction residual:
\begin{equation}\label{eq:error-wr}
\begin{aligned}
    \mathcal E_w
    \leq
    \sum_{k=1}^n
    \mathbb E_{\mathbf x}
    \left[g_k\|\mathbf x\|^2\right]
    \|\mathbf W^k-\hat{\mathbf W}^k\|_F^2.
\end{aligned}
\end{equation}
An analogous bound holds for SwiGLU FFN experts (Appendix~\ref{deri:swiglu}), with a prefactor determined by the original weight norms and the Lipschitz constant of the activation.

\noindent\textit{Remark.} In contrast to the preceding interventions, this bound contains no additional structural term induced by removing experts or tying their routing. The error is instead controlled by the quality of parameter reconstruction.

\paragraph{Regime in modern MoE-based LLMs.} 
The structural terms in \cref{eq:error-irr-m,eq:error-irr-p} are particularly relevant when routing is input-dependent and experts are functionally heterogeneous. Modern MoEs encourage broad expert utilization through load-balancing mechanisms~\citep{lepikhin2021gshard,wang2024auxiliary}, while prior studies report substantial functional specialization among trained experts~\citep{li2025moesvd,lasby2026reap}. These properties suggest that the structural costs identified by the pairwise analysis need not be negligible in pretrained MoEs. We examine this empirically.

\subsection{Empirical evidence for three interventions}
Next, we connect our derivations to empirical observations on two MoE-based LLMs: Qwen3-30B-A3B-Instruct-2507 ($n=128$, top-8; high granularity) and Mixtral-8x7B-v0.1 ($n=8$, top-2; low granularity). We collect activations from 128 sequences × 2048 tokens from the C4 corpus~\cite{raffel2020c4}.

\paragraph{Structural costs are empirically non-trivial.}
We first verify that the structural costs in \cref{eq:error-irr-m,eq:error-irr-p} are non-trivial in pretrained MoEs. \cref{fig:irr-grid} reports $\bar{\mathcal{E}}_p^{\mathrm{irr}}$ and $\bar{\mathcal{E}}_m^{\mathrm{irr}}$ per MoE layer on the original models. We observe that (i) both quantities are strictly above zero across every MoE layer in both architectures, showing that the structural terms are non-vacuous for these models; (ii) both grow $\sim$3--4 orders of magnitude with depth, consistent with expert specialization concentrating in deep layers~\citep{lasby2026reap,bandarkar2026multilingual} --- precisely where compression incurs the highest cost; (iii) within each model the two quantities trace similar shapes across layers, suggesting that both quantities are shaped by common layerwise changes in routing and expert geometry.

\paragraph{Activations drift in compressed models.}
We examine whether the distinctions highlighted by the structural analysis are reflected in realized compression error. To address this, we apply one representative method per family at 25\% MoE compression: REAP~\citep{lasby2026reap} for pruning, HC-SMoE~\citep{chen2025hcsmoe} for merging, and SLBF (\cref{sec:method}) for weight reconstruction. For each compressed model, we measure the per-layer relative MoE-block output discrepancy $\|\delta \mathbf h_\ell\|^2 / \|\mathbf h_\ell^{\mathrm{orig}}\|^2$. \Cref{fig:hidden-layer} reports the results.
SLBF achieves the lowest discrepancy across all layers on both architectures, consistent with its preservation of expert identities and routing. 
The relative ordering of REAP and HC-SMoE changes with expert granularity. On Qwen3, HC-SMoE exhibits larger drift, consistent with the difficulty of tying many distinct experts to shared representatives, consistent with prior observations~\citep{lasby2026reap}; on Mixtral, REAP exhibits larger drift, where the smaller co-active expert set offers fewer substitutes for a removed expert. SLBF remains below both across the two regimes.
\begin{figure*}[t]
    \centering
    \begin{subfigure}[t]{0.24\linewidth}
        \centering
        \vspace{0pt} 
        \tikzset{every picture/.style={line width=1.25pt}} 

\begin{tikzpicture}[x=0.75pt,y=0.75pt,yscale=-1,xscale=1]

\draw    (103.92,27.75) -- (125.88,27.75) ;
\draw    (48.92,27.75) -- (78.92,27.75) ;
\draw    (9.63,59.75) -- (82.88,59.75) ;
\draw   (23.92,15.67) -- (48.92,15.67) -- (48.92,40.67) -- (23.92,40.67) -- cycle ;
\draw  [color={rgb, 255:red, 157; green, 176; blue, 210 }  ,draw opacity=1 ] (78.92,15.67) -- (103.92,15.67) -- (103.92,40.67) -- (78.92,40.67) -- cycle ;
\draw [color={rgb, 255:red, 157; green, 176; blue, 210 }  ,draw opacity=1 ]   (91.38,40.67) -- (91.38,51) ;
\draw  [color={rgb, 255:red, 157; green, 176; blue, 210 }  ,draw opacity=1 ] (82.88,51.12) -- (99.88,51.12) -- (99.88,66.07) -- (82.88,66.07) -- cycle ;
\draw    (9.13,27.25) -- (23.88,27.25) ;
\draw    (36.13,40.75) -- (36.13,59.5) ;
\draw  [color={rgb, 255:red, 0; green, 0; blue, 0 }  ,draw opacity=0.38 ] (70.13,19.7) .. controls (70.13,13.79) and (74.92,9) .. (80.83,9) -- (100.93,9) .. controls (106.83,9) and (111.63,13.79) .. (111.63,19.7) -- (111.63,59.8) .. controls (111.63,65.71) and (106.83,70.5) .. (100.93,70.5) -- (80.83,70.5) .. controls (74.92,70.5) and (70.13,65.71) .. (70.13,59.8) -- cycle ;

\draw (28,20.5) node [anchor=north west][inner sep=0.75pt]    {$\mathbf{A}$};
\draw (83,20.5) node [anchor=north west][inner sep=0.75pt]  [color={rgb, 255:red, 157; green, 176; blue, 210 }  ,opacity=1 ]  {$\mathbf{{\textstyle B}}$};
\draw (83,54) node [anchor=north west][inner sep=0.75pt]  [color={rgb, 255:red, 157; green, 176; blue, 210 }  ,opacity=1 ]  {$\mathbf{w}$};
\draw (4.25,51) node [anchor=north west][inner sep=0.75pt]  [font=\scriptsize]  {$n$};
\draw (57,18) node [anchor=north west][inner sep=0.75pt]  [font=\scriptsize]  {$r$};
\draw (112,13.5) node [anchor=north west][inner sep=0.75pt]  [font=\scriptsize]  {$d_{2}$};
\draw (0,13.5) node [anchor=north west][inner sep=0.75pt]  [font=\scriptsize]  {$d_{1}$};
\draw (110.5,60.4) node [anchor=north west][inner sep=0.75pt]  [font=\scriptsize]  {$\textcolor[rgb]{0.61,0.61,0.61}{\Phi( \cdot )}$};
\draw (93.25,42.15) node [anchor=north west][inner sep=0.75pt]  [font=\scriptsize,color={rgb, 255:red, 157; green, 176; blue, 210 }  ,opacity=1 ]  {$m$};

\end{tikzpicture}
        \par\vspace{1em} 
        \tikzset{every picture/.style={line width=1.25pt}} 

\begin{tikzpicture}[x=0.75pt,y=0.75pt,yscale=-1,xscale=1]

\draw    (112.38,27.75) -- (125.88,27.75) ;
\draw    (49,27.75) -- (69.13,27.75) ;
\draw    (9.63,59.75) -- (82.88,59.75) ;
\draw   (23.92,15.67) -- (48.92,15.67) -- (48.92,40.67) -- (23.92,40.67) -- cycle ;
\draw  [color={rgb, 255:red, 255; green, 130; blue, 93 }  ,draw opacity=1 ] (69.38,15.67) -- (85.38,15.67) -- (85.38,40.67) -- (69.38,40.67) -- cycle ;
\draw [color={rgb, 255:red, 255; green, 130; blue, 93 }  ,draw opacity=1 ]   (85,27.75) -- (96.38,27.75) ;
\draw [color={rgb, 255:red, 255; green, 130; blue, 93 }  ,draw opacity=1 ]   (91.38,46.75) -- (91.38,51) ;
\draw  [color={rgb, 255:red, 255; green, 130; blue, 93 }  ,draw opacity=1 ] (91.38,46.75) -- (77.13,46.75) -- (77.13,40.5) ;
\draw  [color={rgb, 255:red, 255; green, 130; blue, 93 }  ,draw opacity=1 ] (91.38,46.75) -- (104.88,46.75) -- (104.88,40.5) ;
\draw  [color={rgb, 255:red, 255; green, 130; blue, 93 }  ,draw opacity=1 ] (82.88,51.12) -- (99.88,51.12) -- (99.88,66.07) -- (82.88,66.07) -- cycle ;
\draw    (9.13,27.25) -- (23.88,27.25) ;
\draw    (36.13,40.75) -- (36.13,59.5) ;
\draw  [color={rgb, 255:red, 0; green, 0; blue, 0 }  ,draw opacity=0.38 ] (64.13,19.7) .. controls (64.13,13.79) and (68.92,9) .. (74.83,9) -- (106.93,9) .. controls (112.83,9) and (117.63,13.79) .. (117.63,19.7) -- (117.63,59.8) .. controls (117.63,65.71) and (112.83,70.5) .. (106.93,70.5) -- (74.83,70.5) .. controls (68.92,70.5) and (64.13,65.71) .. (64.13,59.8) -- cycle ;
\draw  [color={rgb, 255:red, 255; green, 130; blue, 93 }  ,draw opacity=1 ] (96.38,15.67) -- (112.38,15.67) -- (112.38,40.67) -- (96.38,40.67) -- cycle ;

\draw (28,20.5) node [anchor=north west][inner sep=0.75pt]    {$\mathbf{A}$};
\draw (69,20.5) node [anchor=north west][inner sep=0.75pt]  [color={rgb, 255:red, 255; green, 130; blue, 93 }  ,opacity=1 ]  {$\mathbf{{\textstyle U}}$};
\draw (96,20.5) node [anchor=north west][inner sep=0.75pt]  [color={rgb, 255:red, 255; green, 130; blue, 93 }  ,opacity=1 ]  {$\mathbf{V}$};
\draw (83,54) node [anchor=north west][inner sep=0.75pt]  [color={rgb, 255:red, 255; green, 130; blue, 93 }  ,opacity=1 ]  {$\mathbf{w}$};
\draw (4.25,51) node [anchor=north west][inner sep=0.75pt]  [font=\scriptsize]  {$n$};
\draw (52,18) node [anchor=north west][inner sep=0.75pt]  [font=\scriptsize]  {$r$};
\draw (118,13.5) node [anchor=north west][inner sep=0.75pt]  [font=\scriptsize]  {$d_{2}$};
\draw (0,13.5) node [anchor=north west][inner sep=0.75pt]  [font=\scriptsize]  {$d_{1}$};
\draw (116.5,60.4) node [anchor=north west][inner sep=0.75pt]  [font=\scriptsize]  {$\textcolor[rgb]{0.61,0.61,0.61}{\Phi( \cdot )}$};
\draw (86,15) node [anchor=north west][inner sep=0.75pt]  [font=\scriptsize,color={rgb, 255:red, 255; green, 130; blue, 93 }  ,opacity=1 ]  {$k$};
\draw (105.25,41.15) node [anchor=north west][inner sep=0.75pt]  [font=\scriptsize,color={rgb, 255:red, 255; green, 130; blue, 93 }  ,opacity=1 ]  {$m$};

\end{tikzpicture}
        \par\vspace{1em}
        \caption{Tensor diagrams.}
        \label{fig:tensor-diagram}
    \end{subfigure}
    \hfill
    \begin{subfigure}[t]{0.37\linewidth}
        \centering
        \vspace{0pt}
        \includegraphics[width=\linewidth]{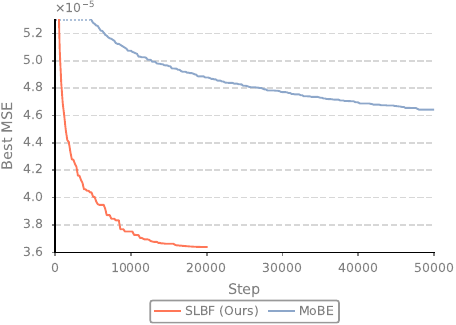}
        \caption{Convergence Profile.}
        \label{fig:slbf-convergence}
    \end{subfigure}
    \hfill
    \begin{subfigure}[t]{0.37\linewidth}
        \centering
        \vspace{0pt}
        \includegraphics[width=\linewidth]{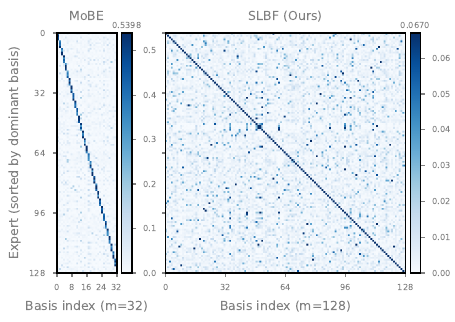}
        \caption{Mixing weight heatmaps.}
        \label{fig:slbf-mixing}
    \end{subfigure}

    \caption{\textbf{SLBF vs.\ MoBE on an MoE layer.} \textbf{(a)} Tensor diagrams of the MoBE (blue, top) and SLBF (orange, bottom) parameterizations; nodes are tensors, edges are indices, and shared edges denote contraction. \textbf{(b)} Reconstruction loss during training on Qwen3-30B-A3B, layer 7, \texttt{gate}: SLBF reaches MoBE's loss in $\sim$2k iterations and improves by $>$20\% by 20k iterations, despite running for fewer steps. \textbf{(c)} Mixing weight $w_{ij}$ heatmaps under the same compression budget: SLBF's lower per-basis cost permits $m = n$, revealing rich cross-expert sharing, while MoBE is restricted to $m < n$ and concentrates each expert on a single dominant basis.}
    \label{fig:main}
\end{figure*}

\section{Method}
\label{sec:method}

The structural analysis in \Cref{sec:cost-analysis} shows that weight reconstruction avoids the intervention-induced structural costs of pruning and merging by preserving expert identities and routing, leaving its compression error controlled by parameter reconstruction quality. This makes parameterization efficiency central to effective weight reconstruction. We therefore address a bottleneck in the state-of-the-art mixture-of-basis-expert formulation (MoBE)~\citep{chen2026mobe}.

\paragraph{MoBE representation.} 
MoBE operates on the projection weights $\mathbf{W}^i\in\{\mathbf W^{i}_{\text{gate}},\mathbf W^{i}_{\text{up}}\}$.\footnote{We write each weight as $\mathbf W^i \in \mathbb R^{d_1 \times d_2}$ 
with $d_1 \leq d_2$, transposing where necessary to fit this convention.} MoBE approximates each expert weight as
\begin{equation}
    \hat{\mathbf W}^i_{\text{MoBE}} = \mathbf A^i\, \Phi\!\left(\sum_{j=1}^{m} w_{ij}\, \mathbf B_j\right),
\end{equation}
where $\{\mathbf B_j\}_{j=1}^m \subset \mathbb R^{r \times d_2}$ is a shared 
basis set, $w_{ij} \in \mathbb R$ is a per-expert mixing coefficient, $\mathbf A^i \in \mathbb R^{d_1 \times r}$ is a per-expert linear adapter, and $\Phi$ is a non-linearity to enhance the representational power. The parameter cost per basis is $r \cdot d_2$, restricting $m$ to be small under tight compression budgets.
MoBE converts trained MoE-based LLM simply through minimizing weight reconstruction loss:
\begin{equation}
    \begin{aligned}
        &\min_{\mathbf w,\{\mathbf{A}^i\}_{i=1}^n,\{\mathbf B_j\}_{j=1}^m } \sum_{i=1}^n\|\mathbf W^i - \hat{\mathbf W}^i\|_F^2 \\
        &= \sum_{i=1}^n\left\|\mathbf W^i - \mathbf{A}^if\left(\sum_{j=1}^m w_{ij} \mathbf{B}_{j}\right)\right\|_F^2.
    \end{aligned}
\end{equation}

\subsection{Improved Weight Reconstruction with Shared Low-rank Basis Factorization}

We propose \textbf{S}hared \textbf{L}ow-rank \textbf{B}asis \textbf{F}actorization (SLBF), which replaces each full basis $\mathbf B_j$ with a rank-$k$ factorization 
$\mathbf U_j \mathbf V_j$, where $\mathbf U_j \in \mathbb R^{r \times k}$ and 
$\mathbf V_j \in \mathbb R^{k \times d_2}$:
\begin{equation}\label{eq:SLBF}
    \hat{\mathbf W}^i = \mathbf A^i\, \Phi\!\left(\sum_{j=1}^{m} w_{ij}\, \mathbf U_j \mathbf V_j\right),
\end{equation}
where $\{\mathbf U_j \mathbf V_j\}_{j=1}^m$ are $m$ shared low-rank bases for reconstructing the right vectors of $\mathbf W^i$, and $\mathbf U_j \in \mathbb R^{r\times k}, \mathbf V_j \in \mathbb R^{k\times d_2}$. Under a fixed compression budget, this factorization affords substantially more bases. \Cref{fig:tensor-diagram} shows the tensor diagram of the proposed factorization. We discuss the structural and empirical implications of this trade-off.

\paragraph{Low-rank mixtures are high-rank.}
Each individual basis $\mathbf U_j \mathbf V_j$ has rank at most $k$. The pre-activation mixture $\sum_{j=1}^{m} w_{ij} \mathbf U_j \mathbf V_j$ generically attains rank $\min(mk, r, d_2)$, matching MoBE's per-basis rank when $mk \geq r$. SLBF thus trades few-rich bases for many-simple bases composed through the mixture, with $\Phi$ providing further non-linear composition.

\begin{figure}
    \centering
    \includegraphics[width=0.9\linewidth]{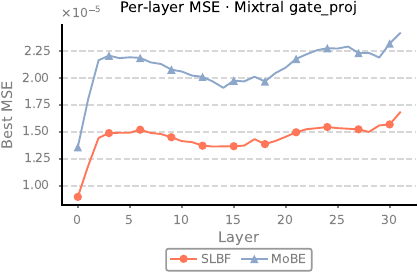}
    \caption{Per-layer MSE on \texttt{Mixtral-8x7B} \texttt{gate}. SLBF achieves uniformly lower weight reconstruction error across all layers under a matched compression.}
    \label{fig:mixtra-gate-recon-loss}
\end{figure}

\paragraph{Empirical benefit of richer mixing.} Empirically, SLBF exploits this freedom by distributing reconstruction across multiple bases per expert, whereas MoBE tends to concentrate each expert on one dominant basis (\cref{fig:slbf-mixing}). SLBF also converges faster and reaches lower reconstruction MSE under fixed budget (\cref{fig:slbf-convergence}). On Mixtral-8x7B, this advantage holds uniformly: SLBF achieves lower reconstruction MSE across all 32 MoE layers under matched budget (\cref{fig:mixtra-gate-recon-loss}).

\paragraph{Connections to block-term decomposition.}
In tensor decomposition terms, the pre-activation mixture $\sum_{j=1}^m w_{ij}\, \mathbf U_j \mathbf V_j$ corresponds to the $i$-th matrix in a rank-$(k, k, 1)$ block-term decomposition (BTD)~\citep{de2008btd} across experts. SLBF thus generalizes MoBE in its basis structure: MoBE's mixture $\sum_j w_{ij} \mathbf B_j$ is the Tucker-1 case recovered at $k = r$, and the rank-$k$ restriction enables the favorable parameter trade-off.
BTD has been applied effectively in model compression~\citep{ye2018btd-rnn,ma2019tensorizedtransformer} and signal processing~\citep{giampouras2022btd,min2026hsi-btd}, motivating its use in our formulation.

\paragraph{Training.}
We fit SLBF by minimizing the weight-space reconstruction loss 
$\sum_i \|\mathbf W^i - \hat{\mathbf W}^i\|_F^2$ independently for each MoE layer. This objective directly reduces the parameter reconstruction residual appearing in~\Cref{eq:error-wr} while requiring no calibration data. The objective is non-convex and solved with AdamW~\citep{loshchilov2018adamw}. For stability, we use techniques including matrix-sign initialization (details in \cref{sec:slbf-config,appendix:training}).

\subsection{Parameter Reduction via Bilinear Symmetry Gauge Fixing}\label{sec:gauge-fixing}

The bilinear product $\mathbf{U}\mathbf{V}$ exhibits a known gauge symmetry~\citep{zhao2025pivoting,lu2026skipcat}: for any invertible 
$\mathbf{Q} \in \mathrm{GL}(k)$, the factorization 
$(\mathbf{U}\mathbf{Q})(\mathbf{Q}^{-1}\mathbf{V}) = \mathbf{U}\mathbf{V}$ is invariant. This leaves $k^2$ parameters redundant in each basis, which we eliminate by fixing the gauge.
Specifically, we can write the basis as:
\begin{equation}
        \mathbf U \mathbf V = \begin{bmatrix} \mathbf U_a \\ \mathbf U_b \end{bmatrix} \mathbf V 
        = \begin{bmatrix} \mathbf U_a\mathbf V \\ \mathbf U_b \mathbf V \end{bmatrix} 
        = \begin{bmatrix}  I_k \\ \mathbf U_b \mathbf U_a^{-1} \end{bmatrix} \mathbf U_a \mathbf V,
\end{equation}
where $\mathbf U_a\in \mathbb R^{k\times k}$ is an invertible matrix.
Assuming that $\mathbf U\in\mathbb R^{r\times k}$ has full column rank, such an invertible $k\times k$ row submatrix always exists. This condition held for all trained factors in our experiments.
Therefore, re-parameterizing $\hat{\mathbf{U}} \leftarrow 
[I_k; \mathbf{U}_b \mathbf{U}_a^{-1}]$ and $\hat{\mathbf{V}} \leftarrow \mathbf{U}_a \mathbf{V}$ preserves the product while the fixed identity block in $\hat{\mathbf{U}}$ removes $k^2$ parameters per basis.
We select the rows forming $\mathbf{U}_a$ via column-pivoted QR on $\mathbf{U}^\top$ for numerical stability.

\paragraph{Parameter count.}
We summarize the parameter count of one projection in an MoE layer in~\Cref{tab:param-count} for MoBE and the proposed SLBF.
The per-expert $\mathbf A^i \in \mathbb R^{d_1 \times r}$ contributes $n d_1 r$ parameters, identical across MoBE and SLBF. The shared-basis cost differs: $m r d_2$ in MoBE versus $m k (r + d_2)$ in SLBF, a ratio of $k(r + d_2)/(r d_2)$. For $r \leq d_2$, SLBF affords proportionally more bases under a fixed compression budget. The gauge-fixing scheme of \cref{sec:gauge-fixing} further removes $m k^2$ parameters from $\hat{\mathbf U}_j$ at no representational cost (ablated in~\cref{app:gauge_ablation},~\Cref{fig:gauge_ablation}).

\begin{table}[]
    \centering
    \small
    \begin{tabular}{ll}
    \toprule
    Method  & \# Parameters \\
    \midrule
    Standard & $n d_1 d_2$ \\
    Pruning/Merging & $n' d_1 d_2$ \\
    MoBE & $n d_1 r + m r d_2 + n m$ \\
    SLBF & $n d_1 r + m k(r + d_2-k) + n m$ \\
    \bottomrule
    \end{tabular}
    \caption{Parameter counts per projection weight. }
    \label{tab:param-count}
\end{table}

\section{Experiments}
\label{sec:experiments}

In this section, we conduct comprehensive experiments on five state-of-the-art MoE-based LLMs, spanning 16B to 122B parameters and diverse architectures, and compare SLBF against representative baselines.

\begin{table*}[t]
\centering
\footnotesize 
\setlength{\tabcolsep}{2.5pt} 
\begin{tabular}{w{c}{1.5cm}|w{c}{0.1cm}|ccc|ccc|ccc|c} 
\toprule
\multirow{2}{*}{\textbf{Method}} & \multirow{2}{*}{\textbf{Ratio}} & \multicolumn{3}{c|}{\textbf{Reasoning ($\uparrow$)}} & \multicolumn{3}{c|}{\textbf{Knowledge ($\uparrow$)}} & \multicolumn{3}{c|}{\textbf{Math/Coding ($\uparrow$)}} & \multirow{2}{*}{\textbf{Avg.($\uparrow$)}} \\
 & & \textbf{ARC-C} & \textbf{GPQA-D} & \textbf{IFEval} & \textbf{MMLU} & \textbf{CEval} & \textbf{CMMLU} & \textbf{GSM8K} & \textbf{MBPP} & \textbf{HEval} & \\
\midrule

\multicolumn{2}{c|}{\textbf{Qwen3-30B-A3B-2507}} & 95.0 & 58.6 & 82.8 & 87.5 & 86.3 & 84.4 & 95.9 & 79.6 & 93.9 & 84.9 \\
\midrule[0.2pt]
REAP & 24\% & 88.4 & 36.4 & 80.6 & 75.7 & 65.7 & 60.9 & 95.5 & 67.4 & 82.3 & 72.5 \\
HC-SMoE & 24\%  & 93.9 & 52.5 & 81.9 & 84.1 & 69.3 & 67.5 & 94.7 & 44.0 & 79.9 & 74.2 \\
REAM & 24\% & 89.9 & 39.9 & 81.2 & 78.5 & 65.7 & 58.8 & 94.7 & 66.8 & 86.6 & 73.6 \\
MoLAE  & 24\% & 91.3 & 46.5 & 77.1 & 84.3 & 73.8 & 73.3 & 89.3 & 64.6 & 84.2 & 76.0 \\
MoBE & 24\% & \textbf{94.0} & 54.6 & 82.4 & 85.5 & 82.8 & 81.6 & 95.2 & 77.4 & 91.5 & 82.8 \\
\textbf{SLBF} & 24\% & 93.9 & \textbf{60.1} & \textbf{82.6} & \textbf{85.9} & \textbf{85.1} & \textbf{82.2} & \textbf{95.6} & \textbf{78.6} & \textbf{95.1} & \textbf{84.3} \\
\cmidrule(lr){1-12}
MoLAE  & 32\% & 84.5 & 34.3 & 69.3 & 75.8 & 66.3 & 66.4 & 80.0 & 51.0 & 67.7 & 66.1 \\
MoBE & 32\% & 91.6 & 46.5 & 79.9 & 82.3 & 77.1 & 76.2 & 92.9 & 69.6 & 89.0 & 78.3 \\
\textbf{SLBF} & 36\% & \textbf{93.0} & \textbf{57.6} & \textbf{83.9} & \textbf{85.5} & \textbf{80.5} & \textbf{79.9} & \textbf{94.7} & \textbf{76.2} & \textbf{90.9} & \textbf{82.4} \\

\midrule

\multicolumn{2}{c|}{\footnotesize{\textbf{Moonlight-16B-A3B}}} &  83.6 & 31.8 & 43.4 & 71.7 & 71.4 & 73.4 & 74.8 & 57.4 & 71.3 & 64.3 \\
\midrule[0.2pt]
MoLAE & 15\% &  65.1 & 18.2 & 20.9 & 48.9 & 50.7 & 52.6 & 16.1 & 10.8 & 0.0 & 31.5 \\
MoBE & 15\% & 79.1 & 23.7 & 35.5 & 61.6 & 65.2 & 66.8 & 52.2 & 46.0 & 59.8 & 54.4 \\
\textbf{SLBF} & 15\%  & \textbf{80.8} & \textbf{28.8} & \textbf{40.7} & \textbf{69.2} & \textbf{69.0} & \textbf{71.9} & \textbf{71.7} & \textbf{59.6} & \textbf{70.7} & \textbf{62.5} \\

\midrule

\multicolumn{2}{c|}{\footnotesize{\textbf{Gemma4-26B-A4B-it}}}   & 94.1 & 65.7 & 90.4 & 89.3 & 69.3 & 73.1 & 94.7 & 91.2 & 98.2 & 85.1 \\
\midrule[0.2pt]
MoLAE & 37\%  & 74.2 & 29.8 & 71.2 & 67.0 & 27.7 & 51.0 & 70.7 & 7.8 & 33.5 & 48.1 \\
MoBE & 37\%   & 91.8 & 46.0 & 85.6 & 82.1 & 28.1 & 63.4 & 88.5 & 35.2 & 86.6 & 67.5 \\
\textbf{SLBF} & 37\%  & \textbf{92.9} & \textbf{61.6} & \textbf{89.3} & \textbf{87.8} & \textbf{63.6} & \textbf{68.6} & \textbf{93.6} & \textbf{84.6} & \textbf{93.9} & \textbf{81.8} \\

\midrule

\multicolumn{2}{c|}{\footnotesize{\textbf{Qwen3.5-122B-A10B}}}  & 96.1 & 83.3 & 93.0 & 88.7 & 86.7 & 86.2 & 97.4 & 53.0 & 92.1 & 86.3 \\
\midrule[0.2pt]
MoLAE & 34\%    & 93.3 & 43.4 & 70.4 & 79.5 & 78.6 & 79.4 & 91.3 & 38.8 & 76.2 & 72.3 \\
MoBE & 34\%   & 96.2 & 67.7 & 87.8 & 87.7 & 84.0 & 84.4 & 97.0 & 47.0 & 93.3 & 82.8 \\
\textbf{SLBF} & 34\%  & \textbf{96.3} & \textbf{83.8} & \textbf{93.7} & \textbf{89.6} & \textbf{87.4} & \textbf{86.9} & \textbf{97.4} & \textbf{61.8} & \textbf{94.5} & \textbf{88.0} \\

\bottomrule
\end{tabular}
\caption{Multi-architecture compression results. We report nine benchmarks grouped into reasoning, knowledge, and math/coding for four instruction-tuned MoE LLMs at varying compression ratios. All methods are applied one-shot without fine-tuning. Ratio denotes total parameter reduction. Best results per model and ratio are \textbf{bolded}.}
\label{tab:main1}
\end{table*}
\begin{table*}[t]
\centering
\footnotesize 
\setlength{\tabcolsep}{2.5pt} 
\begin{tabular}{w{c}{1.5cm}|w{c}{0.1cm}|c|ccccccc|c} 
\toprule
 \textbf{Method} & \textbf{Ratio} & \textbf{WT ($\downarrow$)} & \textbf{ARC-C} & \textbf{ARC-E} & \textbf{HellaSwag} & \textbf{OPQA} & \textbf{RTE} & \textbf{Winogrande} & \textbf{PIQA} & \textbf{Avg.($\uparrow$)} \\
\midrule
\multicolumn{2}{c|}{\footnotesize{\textbf{Mixtral-8x7B-v0.1}}} & 5.64 & 59.7 & 82.7 & 84.2 & 49.4 & 71.1 & 77.3 & 83.2 & 72.5 \\
\midrule[0.2pt]
Frequency & 24\% & 9.10 & 44.8 & 70.3 & 66.6 & 30.2 & 58.5 & 56.8 & 77.6 & 57.8 \\
MC-SMoE & 24\% & 48.42 & 29.4 & 52.0 & 49.4 & 30.6 & 52.7 & 54.1 & 66.2 & 47.8 \\
HC-SMoE & 24\% & 8.54 & 53.2 & 78.1 & 81.3 & 46.2 & 67.9 & 75.2 & 81.7 & 69.1 \\
EAN & 24\% & 8.28 & 54.5 & \textbf{80.6} & 81.9 & 45.4 & \textbf{69.7} & 75.9 & \textbf{83.1} & 70.2 \\
REAP & 24\% & 8.30 & 53.9 & 79.5 & 81.8 & 46.8 & 67.2 & \textbf{77.1} & 82.5 & 69.8 \\
Sub-MoE & 24\% & 17.51 & 43.0 & 66.5 & 68.0 & 35.8 & 53.1 & 68.0 & 74.0 & 58.3 \\
TD-MoE & 24\% & 16.68 & 42.5 & 69.3 & 66.7 & 40.4 & 55.6 & 69.4 & 74.8 & 59.8 \\
\textbf{SLBF} & 24\% & \textbf{6.92} & \textbf{56.7} & \textbf{80.6} & \textbf{83.5} & \textbf{50.4} & 63.9 & 76.4 & 83.0 & \textbf{70.7} \\
\cmidrule(lr){1-11}
MoLAE & 30\% & 11.17 & 51.6 & 76.6 & 77.9 & 46.0 & 58.1 & 75.4 & 80.8 & 66.6 \\
MoBE & 30\% & 8.59 & 54.0 & 79.8 & 80.8 & \textbf{49.4} & 62.1 & 75.6 & 82.3 & 69.1 \\
\textbf{SLBF} & 30\% & \textbf{7.78} & \textbf{55.8} & \textbf{80.1} & \textbf{82.1} & 48.6 & \textbf{67.5} & \textbf{76.7} & \textbf{83.1} & \textbf{70.6} \\
\bottomrule
\end{tabular}
\caption{Compression comparison on Mixtral-8x7B-v0.1. We report WikiText-2 perplexity (WT ($\downarrow$)) and zero-shot task accuracy. All methods are applied one-shot without fine-tuning. Best results per compression ratio are \textbf{bolded}.}
\label{tab:main2}
\end{table*}

\subsection{Setup}

\paragraph{Models.}
We evaluate SLBF on five MoE LLMs spanning a range of architectures, granularities, and scales: Moonlight-16B-A3B-Instruct~\citep{liu2025moonlight}, Qwen3-30B-A3B-Instruct-2507~\citep{qwen3technicalreport}, Gemma4-26B-A4B-it~\cite{gemma4}, Mixtral-8x7B-v0.1 (base)~\cite{jiang2024mixtral}, Qwen3.5-122B-A10B~\cite{qwen3.5}. Compression ratios (\textbf{Ratio}) refer to parameter reduction with respect to the total model.

\paragraph{Baselines.}
We compare against representative methods from all three families. \emph{Pruning}: Frequency, REAP~\citep{lasby2026reap}, and EAN~\citep{jaiswal2025ean}, each using a different expert-saliency criterion. \emph{Merging}: MC-SMoE~\citep{li2024mcsmoe}, HC-SMoE~\citep{chen2025hcsmoe}, and REAM~\citep{jha2026ream}. \emph{Weight reconstruction}: MoLAE~\citep{liu2025molae}, and MoBE~\citep{chen2026mobe}. \emph{Others}: Sub-MoE~\citep{li2026submoe}, TD-MoE~\citep{xu2026tdmoe}. MoLAE, MoBE, and SLBF are \textit{data-free}, while the remaining baselines use calibration data. For all calibration-based baselines, we use 1024 C4 sequences of 2048 tokens.

\paragraph{Evaluation benchmarks.}
We evaluate on a comprehensive set of benchmark datasets. 
For the modern instruction-tuned MoE LLMs (\cref{sec:main-results}), we report nine standard benchmarks grouped as \textit{reasoning}: ARC-Challenge~\citep{clark2018arc}, GPQA-Diamond~\citep{rein2024gpqa}, IFEval~\citep{zhou2023ifeval}; \textit{knowledge}: MMLU~\citep{hendrycks2021mmlu}, CEval~\citep{huang2023ceval}, CMMLU~\citep{li2024cmmlu}; and \textit{math/coding}: GSM8K~\citep{cobbe2021gsm8k}, MBPP~\citep{austin2021mbpp}, HumanEval~\citep{chen2021humaneval}. 
For Mixtral-8x7B, following prior Mixtral compression work, we report perplexity on WikiText-2 and zero-shot accuracy on ARC-Challenge, ARC-Easy~\citep{clark2018arc}, HellaSwag~\citep{zellers2019hellaswag}, OpenBookQA~\citep{mihaylov2018opqa}, RTE~\citep{bentivogli2009rte}, WinoGrande~\citep{sakaguchi2021winogrande}, and PIQA~\citep{2020piqa}.
Using each model's conventional suite preserves direct comparability with prior compression work. Evaluations are conducted using \texttt{lm-evaluation-harness}~\citep{eval-harness}.
Original and compressed checkpoints are evaluated with identical task definitions and decoding configurations.
Full evaluation details, including exact task IDs, prompting settings, and decoding settings, are provided in \Cref{app:evaluation}.

\paragraph{SLBF configuration.}
\label{sec:slbf-config}
We use $m=n$ shared bases throughout, matching the number of bases to the number of experts and yielding an $n\times n$ expert-to-basis mixing structure (\cref{fig:slbf-mixing}). We set $r=d_1$ following~\citet{chen2026mobe}, so no low-rank bottleneck is introduced in $\mathbf{A}^i$; the per-basis rank $k$ is chosen to match the target compression ratio. 
We use the SiLU activation~\citep{elfwing2018silu} for $\Phi$. We optimize with AdamW~\cite{loshchilov2018adamw}. Because of the multiplicative low-rank parameterization, we use separate learning-rate groups for $\{\mathbf A,\mathbf w\}$ and $\{\mathbf U,\mathbf V\}$, together with matrix-sign initialization. Details and remaining hyper-parameters can be found in~\cref{appendix:training}.

\subsection{Multi-architecture Results}
\label{sec:main-results}

We evaluate SLBF on Qwen3-30B-A3B-Instruct-2507, Qwen3.5-122B-A10B, Moonlight-16B-A3B, and Gemma4-26B-A4B-it. Baselines are the strongest representative from each compression family evaluated at this scale: REAP (pruning), HC-SMoE (merging), REAM (merging), MoLAE (weight reconstruction), and MoBE (weight reconstruction). All results are obtained in the one-shot setting with no fine-tuning after compression. 

\paragraph{Performance across models and compression ratios.}
\cref{tab:main1} shows that SLBF consistently outperforms baselines across four architectures and varying compression ratios. At 24\% compression, SLBF preserves near-original accuracy on Qwen3-30B while the strongest baseline MoBE drops 2.1 points and pruning/merging baselines drop over 8 points. The advantage widens at higher compression and on other architectures, reaching a 14.3-point gap over MoBE on Gemma4-26B and a 5.2-point gap on Qwen3.5-122B, where SLBF maintains performance comparable to the original model under the same evaluation configuration. Notably, SLBF at 36\% compression on Qwen3-30B still outperforms MoBE at the easier 32\% ratio. 

\paragraph{Performance on Moonlight.}
Moonlight-16B-A3B differs substantially from the other evaluated models in its training recipe, notably being trained end-to-end with the Muon optimizer~\citep{liu2025moonlight}. SLBF remains effective in this setting: at 15\% compression, the average score drops by only 1.8 points, compared with 9.9 points for MoBE and over 30 points for MoLAE.

\subsection{Cross-family Comparison on Mixtral}
\label{sec:mixtral-results}

Mixtral is more often used in MoE compression literature, on which we compare against a richer set of baselines, including Frequency, EAN~\citep{jaiswal2025ean}, MC-SMoE~\citep{li2024mcsmoe}, HC-SMoE~\citep{chen2025hcsmoe}, Sub-MoE~\citep{li2026submoe}, TD-MoE~\citep{xu2026tdmoe}, MoLAE~\citep{liu2025molae}, and MoBE~\citep{chen2026mobe}. \cref{tab:main2} reports the results.

\paragraph{Performance across compression families.}
At 24\% compression, SLBF reaches 70.7 average accuracy and WikiText-2 perplexity of 6.92, outperforming the strongest pruning baseline (EAN: 70.2 / 8.28) and the strongest merging baseline (HC-SMoE: 69.1 / 8.54). The perplexity gap is more pronounced: SLBF is 1.4 lower (16\% relative) than the next-best method, while the accuracy gap remains within 0.5 points, suggesting better preservation of language-modeling performance. At 30\% compression, SLBF also outperforms the weight-reconstruction baselines MoLAE and MoBE. It maintains the lead with perplexity 7.78 against MoBE's 8.59. These results are consistent with the structural distinction highlighted in \Cref{sec:cost-analysis}: weight reconstruction preserves expert identities and routing, while SLBF further improves the parameterization within this family.

\subsection{Ablation study and runtime analysis}
\label{sec:ablation}

We ablate design choices and analyze runtime cost; full details are in the appendix. SLBF without gauge fixing already outperforms MoBE at every compression ratio; gauge fixing adds a secondary gain (\Cref{app:gauge_ablation}). Compressing all three projections outperforms two at matched budget (62.5 vs 58.2 on Moonlight-16B; \Cref{app:downstream_ablation}), and sweeping $m$ at fixed budget suggests that $m=n$ lies near a strong operating region (\Cref{app:m_ablation}). 
On Mixtral gate projections, SLBF takes 40.9 minutes per (layer, projection) pair versus 94.3 minutes for MoBE, using only 40\% as many training iterations due to earlier convergence, and the procedure is embarrassingly parallel across pairs (\Cref{app:runtime}). In runtime-decompressed serving on $2\times$H100, SLBF reduces model-weight memory by 30\% and increases KV-cache capacity and maximum serving concurrency by 53\%, while retaining $0.88\times$ throughput (\Cref{app:inference}).

\section{Conclusion}

We introduced Shared Low-rank Basis Factorization (SLBF), a data-free weight reconstruction method for MoE compression that uses shared rank-$k$ bases to improve parameter efficiency under fixed compression budgets. Our structural analysis shows that pruning and merging can incur intervention-induced costs associated with expert removal and tied routing, whereas weight reconstruction preserves expert identities and routing and leaves its error controlled by parameter reconstruction quality. Across five MoE architectures spanning 16B to 122B parameters, SLBF consistently outperforms representative methods.
\section*{Limitations}

SLBF operates entirely in weight space without calibration data or post-compression fine-tuning. We therefore do not explore calibration-guided budget allocation or fine-tuning to recover residual accuracy loss; efficient forward computation through the factorized representation is also beyond the scope of this work.

Our study focuses on expert FFN weight compression. Interactions with other compression axes, such as quantization or attention compression, are not explored. We also use a fixed decomposition structure with $r=d_1$ and $m=n$ across models and layers. Although \Cref{app:m_ablation} identifies $m=n$ as a strong operating point, adaptive per-layer or per-projection choices of $(r,k,m)$ may yield further improvements.

Finally, our deployment evaluation covers limited hardware and serving configurations. Runtime-decompressed inference exhibits a memory--throughput trade-off that may differ with optimized MoE kernels or other serving systems. Direct loading of the factorized representation can also incur substantial reconstruction overhead for large MoEs in our current implementation.

\section*{Acknowledgements}
This work was supported by JSPS KAKENHI Grant Numbers JP26K02984 and JP23K28109, and JSPS Bilateral Program Number JPJSBP120257420.

\bibliography{ref}

\appendix

\section{Appendix}
\label{sec:appendix}

\subsection{Missing Derivations in Sec.~\ref{sec:cost-analysis}}
\label{appendix:derivations}
\paragraph{Derivations of merging error (following~\citet{lasby2026reap}).}
Recall the original pair contribution
\begin{equation}
    (g_i+g_j) \cdot (r(\mathbf{x})f^i(\mathbf{x}) +(1-r(\mathbf{x}))f^j(\mathbf{x})).
\end{equation}
Merging replaces by $\tilde f^{ij} = \alpha f_i + (1-\alpha) f_j$ for a constant $\alpha$, a simplification justified in the latest merging method~\citep{chen2025hcsmoe}. The merged contribution is $(g_i+g_j) \tilde f^{ij}$. The output difference of the pair is
\begin{equation}
\begin{aligned}
\delta \mathbf h &= (g_i + g_j) \left(rf^i + (1-r)f^j\right) \\
             &\phantom{{}={}} - (g_i + g_j) \left(\alpha f^i + (1-\alpha)f^j \right) \\
             &= (g_i + g_j) (r-\alpha) (f^i - f^j) \\
             &= (g_i + g_j) (r(\mathbf{x}) - \alpha) \Delta_{ij}(\mathbf{x}).
\end{aligned}
\end{equation}
The expectation of the error squared norm is
\begin{equation}
    \mathcal{E}_m(\alpha) = \mathbb{E}_\mathbf{x}\left[(g_i + g_j)^2 (r(\mathbf{x}) - \alpha)^2 \|\Delta_{ij}(\mathbf{x})\|^2\right].
\end{equation}
The optimal mixing ratio, obtained by setting $\frac{d}{d\alpha} \mathcal{E}_m =0$, is
\begin{equation}
    \alpha^\star = \frac{\mathbb E_{\mathbf x} \left[ (g_i + g_j)^2 \|\Delta_{ij}\|^2 \cdot r(\mathbf x) \right]}{\mathbb E_{\mathbf x} \left[ (g_i + g_j)^2 \|\Delta_{ij}\|^2 \right]}.
\end{equation}
The exact optimal static mixing coefficient is given above. For an interpretable approximation, we additionally assume weak correlation among $(g_i + g_j)^2 $, $\|\Delta_{ij}\|^2$, and $r(\mathbf x)$ across inputs. We refer to the resulting minimum residual within this pairwise static-merging intervention as its irreducible structural cost. The approximate optimal mixing ratio is $\hat{\alpha}^\star = \mathbb E_{\mathbf{x}}[r(\mathbf x)]$. We define the irreducible error as the expectation of the approximate error under $\hat{\alpha}^\star$:
\begin{equation}
    \begin{aligned}
            \mathcal E^{\mathrm{irr}}_m &:= \mathcal E_m(\hat{\alpha}^\star) \\ 
            & \approx \mathbb{E}_{\mathbf x}\left[(g_i + g_j)^2 \right]\cdot \operatorname{Var}\left[r(\mathbf x)\right]\cdot \mathbb{E}_{\mathbf x} \|\Delta_{ij}\|^2.
    \end{aligned}
\end{equation}

\paragraph{Derivations of pruning error.}
Recall the original pair contribution
\begin{equation}
    g_i(\mathbf x) f^i (\mathbf x) + g_j(\mathbf x) f^j (\mathbf x).
\end{equation}
When expert-$j$ is pruned, the output difference is 
\begin{equation}
    \begin{aligned}
        \delta\mathbf h &= g_j f_j - (g'_i - g_i)f_i.
    \end{aligned}
\end{equation}
To wipe out the influence of $g'_i$ choices, we consider the following oracle router after pruning that minimizes the error squared norm
\begin{equation}
    \min_{g'_i}\|g_j f_j - (g'_i - g_i)f_i\|^2.
\end{equation}
The oracle router is $g^\star_i = g_i + \frac{g_j \langle f_j, f_i\rangle}{\|f_i\|^2}$. 
The oracle router depends on $f_j(\mathbf x)$, which can only be computed if expert-$j$ is retained. Since the goal of pruning is to remove expert-$j$ entirely, the dependence renders the oracle structurally unrealizable. Substituting back the oracle router, the expectation of the error squared norm is
\begin{equation}
    \begin{aligned}
        \mathcal E^{\mathrm{irr}}_p &= \mathbb E_\mathbf x\left[\|g_j f_j - (g^\star_i - g_i)f_i\|^2\right]\\
        &= \mathbb E_\mathbf x \left[\|g_j f_j - \frac{g_j \langle f_j, f_i\rangle}{\|f_i\|^2}\cdot f_i\|^2\right]\\
        &=\mathbb E_\mathbf x\Bigg[ g_j^2 \|f_j\|^2 -2\cdot\frac{g_j^2\langle f_j, f_i\rangle}{\|f_i\|^2}\cdot \langle f_j,f_i\rangle \\
        &\phantom{===} + \frac{g_j^2|\langle f_j, f_i\rangle|^2}{\|f_i\|^2} \Bigg]\\
        &=\mathbb E_\mathbf x\left[g^2_j \left(\|f_j\|^2-\frac{|\langle f_j, f_i\rangle|^2}{\|f_i\|^2}\right)\right].
    \end{aligned}
\end{equation}

\paragraph{Derivations of weight reconstruction error bound.}
Each $\mathbf W^k$ is replaced by $\hat {\mathbf W}^k$. Routing and expert presence are preserved.
We start with the linear case $f^k(\mathbf x) = \mathbf x\mathbf W^k$, and consider the influence of all experts. The output difference is
\begin{equation}
    \delta \mathbf h = \mathbf x \, \sum_{k\in\mathcal P(\mathbf x)}g_k  (\mathbf W^k - \hat {\mathbf W}^k).
\end{equation}
We use the fact that $\sum_{k=1}^n g_k=1,\,g_k>0$ and that  $G(\mathbf x):=\sum_{k\in \mathcal P(\mathbf x)} g_k\leq1$. Using Jensen's inequality and Cauchy-Schwarz inequality, we have the squared norm of the difference bounded as
\begin{equation}
\begin{aligned}
    \|\delta \mathbf h\|^2 &= \|\mathbf x  \sum_{k\in\mathcal P(\mathbf x)}g_k  (\mathbf W^k - \hat {\mathbf W}^k)\|^2\\
    &\leq G(\mathbf x) \sum_{k\in \mathcal P(\mathbf x)} g_k \|\mathbf x(\mathbf W^k - \hat {\mathbf W}^k)\|^2 \\
    &\leq  \sum_{k\in \mathcal P(\mathbf x)} g_k \|\mathbf x(\mathbf W^k - \hat {\mathbf W}^k)\|^2 \\
    &\leq  \sum_{k=1}^n g_k \|\mathbf x(\mathbf W^k - \hat {\mathbf W}^k)\|^2. \\
    &\leq \sum_{k=1}^n g_k \|\mathbf x\|^2\|\mathbf W^k - \hat {\mathbf W}^k\|_F^2
\end{aligned}
\end{equation}
Defining $\mathcal E_w :=\mathbb E_\mathbf x \|\delta \mathbf h\|^2$, it is bounded as
\begin{equation}
    \begin{aligned}
        \mathcal E_w\leq \sum_{k=1}^n\|\mathbf W^k - \hat {\mathbf W}^k\|_F^2 \cdot\mathbb E [g_k \|\mathbf x\|^2].
    \end{aligned}
\end{equation}
Modern MoE-based LLMs commonly apply RMSNorm~\citep{zhang2019rms} before the expert FFN. When $\mathbf{x}$ denotes the post-RMSNorm activation and the RMSNorm gain is $\boldsymbol{\gamma}$, we have
\begin{equation}\label{eq:rms}
    \|\mathbf{x}\|^2
    \leq d\|\boldsymbol{\gamma}\|_\infty^2
    \leq C_{\mathrm{RMS}} d,
\end{equation}
where $C_{\mathrm{RMS}}$ is a finite layer-wise bound on the squared
gain magnitude. Therefore,
\begin{equation}
    \mathcal E_w
    \leq
    C_{\mathrm{RMS}} d
    \sum_{k=1}^n
    \mathbb E[g_k]\,
    \|\mathbf W^k-\hat{\mathbf W}^k\|_F^2.
\end{equation}

\paragraph{Weight reconstruction error bound extended to practical SwiGLU-style FFNs.}
\label{deri:swiglu}
Beyond the linear expert, we can extend to the SwiGLU-style FFN case. For SwiGLU FFNs with SiLU activation $\sigma$, the expert is
\begin{equation}
    f^k(\mathbf x) = \left(\sigma(\mathbf x \mathbf W^k_g) \odot \mathbf x \mathbf W^k_u\right) \mathbf W^k_d.
\end{equation}
Compression replaces each of $\mathbf W^k_g, \mathbf W^k_u, \mathbf W^k_d$ with 
$\hat{\mathbf W}^k_g, \hat{\mathbf W}^k_u, \hat{\mathbf W}^k_d$; denote 
$\Delta^k_m := \mathbf W^k_m - \hat{\mathbf W}^k_m$ for $m \in \{u, g, d\}$. 
Define intermediate quantities
\begin{equation}
    \mathbf a^k := \sigma(\mathbf x \mathbf W^k_g), \quad 
    \mathbf b^k := \mathbf x \mathbf W^k_u, \quad 
    \mathbf z^k := \mathbf a^k \odot \mathbf b^k,
\end{equation}
and analogously $\hat{\mathbf a}^k, \hat{\mathbf b}^k, \hat{\mathbf z}^k$ for the perturbed expert. The per-expert error telescopes as
\begin{equation}
\begin{aligned}
    f^k - \hat f^k & = \mathbf z^k \Delta^k_d 
    + \big((\mathbf a^k - \hat{\mathbf a}^k) \odot \mathbf b^k\big)\hat{\mathbf W}^k_d \\
    &\phantom{=} + \big(\hat{\mathbf a}^k \odot \mathbf x \Delta^k_u\big)\hat{\mathbf W}^k_d.
\end{aligned}
\end{equation}
Note that the SiLU activation is $1.1$-Lipchitz (shown in~\citet{sun2025entropybased}, Appendix E).
The triangle inequality, together with $|\sigma(t)| \leq |t|$, $|\sigma(a)-\sigma(b)| \leq L_\sigma |a-b|$, 
the Hadamard inequality $\|\mathbf p \odot \mathbf q\| \leq \|\mathbf p\|_\infty \|\mathbf q\|$, and 
$\|\mathbf x \mathbf M\| \leq \|\mathbf x\| \|\mathbf M\|_F$, yields
\begin{equation}
\begin{aligned}
        &\|f^k - \hat f^k\| \leq \|\mathbf x\|^2 \Big[
        \sqrt{A^k_d}\, \|\Delta^k_d\|_F \\
        &+ \sqrt{A^k_g}\, \|\Delta^k_g\|_F + \sqrt{A^k_u}\, \|\Delta^k_u\|_F
    \Big],
\end{aligned}
\end{equation}
with architectural constants
\begin{equation}
    \begin{aligned}
            &A^k_d := \|\mathbf W^k_g\|_F^2 \|\mathbf W^k_u\|_F^2, \\ 
    &A^k_g := L_\sigma^2 \|\mathbf W^k_u\|_F^2 \|\hat{\mathbf W}^k_d\|_F^2, \\ 
    &A^k_u := \|\hat{\mathbf W}^k_g\|_F^2 \|\hat{\mathbf W}^k_d\|_F^2.
    \end{aligned}
\end{equation}
Using the Cauchy-Schwarz inequality with 
$a_m = \sqrt{A^k_m}$, $b_m = \|\Delta^k_m\|_F$:
\begin{equation}
    \|f^k - \hat f^k\|^2 \leq \|\mathbf x\|^4 \cdot A^k \cdot \|\Delta^k\|_F^2,
\end{equation}
where $A^k := A^k_d + A^k_g + A^k_u$ is a single per-expert architectural constant and 
$\|\Delta^k\|_F^2 := \|\Delta^k_d\|_F^2 + \|\Delta^k_g\|_F^2 + \|\Delta^k_u\|_F^2$ is the total per-expert weight residual. 
Applying the same Jensen argument across routed experts and using~\cref{eq:rms}, we obtain
\begin{equation}
    \mathcal E_w \leq (C_{\mathrm{RMS}} d)^2 \sum_{k}\mathbb E[g_k] A^k \|\Delta^k\|^2_F.
\end{equation}

\begin{table*}[t]
\centering
\small
\begin{tabular}{lrrrrrrrl}
\toprule
Model & $n$ & $d_1\!\times\!d_2$ & Ratio & $k$ & $\eta$ & $\rho$ & iters & projs.\\
\midrule
Mixtral-8x7B-v0.1
& 8 & $4096\!\times\!14336$
& 24\% & 1152 & 0.03 & 0.10 & 8\,000
& $\mathrm{gate},\mathrm{up}$ \\
&
&
& 30\% & 832 & 0.03 & 0.10 & 8\,000
& $\mathrm{gate},\mathrm{up}$ \\
\midrule
Qwen3-30B-A3B-Instruct-2507
& 128 & $768\!\times\!2048$
& 24\% & 226 & 0.10 & 0.30 & 20\,000
& $\mathrm{gate},\mathrm{up},\mathrm{down}$ \\
&
&
& 36\% & 147 & 0.10 & 0.30 & 20\,000
& $\mathrm{gate},\mathrm{up},\mathrm{down}$ \\
\midrule
Qwen3.5-122B-A10B
& 256 & $1024\!\times\!3072$
& 34\% & 250 & 0.10 & 0.222 & 8\,000
& $\mathrm{gate},\mathrm{up},\mathrm{down}$ \\
\midrule
Moonlight-16B-A3B-Instruct
& 64 & $1408\!\times\!2048$
& 15\% & 126 & 0.10 & 0.359 & 20\,000
& $\mathrm{gate},\mathrm{up},\mathrm{down}$ \\
\midrule
Gemma-4-26B-A4B-it
& 128 & $704\!\times\!2816$
& 37\% & 199 & 0.10 & 0.267 & 20\,000
& $\mathrm{gate},\mathrm{up},\mathrm{down}$ \\
\bottomrule
\end{tabular}

\caption{SLBF training hyperparameters for each evaluated compression setting.
$n$ is the number of experts; $d_1\times d_2$ is the per-expert FFN
projection shape; \textbf{Ratio} denotes total-model parameter reduction,
and $k$ is the corresponding per-basis rank. The basis-side parameters
$\{\mathbf A,\mathbf w\}$ use learning rate $\eta$, while
$\{\mathbf U,\mathbf V\}$ use $\rho\eta$.}
\label{tab:slbf-hparams}
\end{table*}

\subsection{SLBF training details}
\label{appendix:training}

All reported results are from a single run; the initialization is deterministic, so results are reproducible given fixed hyperparameters. In our experiments, we train SLBF using one NVIDIA RTX A5000 24GB or NVIDIA RTX A6000 48GB.

\paragraph{Matrix-sign initialization.}
Each expert's basis triple $(\mathbf{A}^{i}, \mathbf{U}_{i}, \mathbf{V}_{i})$ is initialized from the SVD $\mathbf{W}^{i} = \mathbf{U}_W \boldsymbol{\Sigma}_W \mathbf{V}_W$ of the original per-expert weight matrix, with the singular spectrum discarded. Concretely, $\mathbf{A}^{i} := \mathbf{U}_W[:,\!:r]$ is the left singular matrix (top-$r$ columns) and is full-rank since we set $r=d_1$; $\mathbf{U}_{i} := [\,\mathbf{I}_k\,;\,\mathbf{0}\,] \in \mathbb{R}^{r\times k}$ is a partial-identity block (the first $k$ rows form the identity, remaining rows are zero), giving orthonormal columns; and $\mathbf{V}_{i} := \mathbf{V}_W[\,:k,\,:] \in \mathbb{R}^{k\times d_2}$ is the reduced right-singular factor (top-$k$ orthonormal columns). The product $\mathbf{A}^{i} \mathbf{U}_{i} \mathbf{V}_{i}$ therefore lies in the rank-$k$ singular subspace of $\mathbf{W}^{i}$ but carries no magnitude; the optimizer recovers scale and inter-expert mixing during training. We follow~\citep{chen2026mobe} to parameterize $w$ by softmax to keep $\sum_{j=1}^m w_{ij}=1$ for stability. For the mixing weights we set the \emph{pre-softmax} logit $\tilde{w}_{ij} = c\cdot \mathbf{1}[j=i]$ with $c=1$, so the post-softmax weight is a soft (not sharp) one-hot vector: with $n=128$, $\mathrm{softmax}(\tilde w^{i})_e \approx 0.021$ and the remaining $n{-}1$ entries each $\approx 0.0077$. This intentional softness keeps gradients flowing through every basis from step 0; a much sharper initialization strongly suppresses off-diagonal mixing gradients, which we observed to hinder experts from learning to combine bases.

\paragraph{Shared training recipe.}
All SLBF runs use AdamW~\citep{loshchilov2018adamw} with weight decay $2\times 10^{-2}$, $(\beta_1,\beta_2)=(0.9,0.999)$, a cosine learning-rate schedule with $\eta_\text{min}=10^{-4}$, and \texttt{global\_std} normalization of the per-expert reconstruction target. Each $($layer, projection $\in\{\mathrm{gate},\mathrm{up},\mathrm{down}\}$$)$ pair is trained independently; \emph{no calibration data is used}, since the reconstruction objective is purely weight-side. 
Two parameter groups receive separate learning rates: the basis side $\{\mathbf{A}, \mathbf{w}\}$ and the spectrum-restoring side $\{\mathbf{U},\mathbf{V}\}$ at $\eta$ and $\rho\eta$, respectively. Model-specific values of $\eta$ and $\rho$ are listed in Table~\ref{tab:slbf-hparams}.

\paragraph{Projection scope.}
For Qwen3, Qwen3.5, Moonlight, and Gemma-4 we compress all three FFN projections (\texttt{gate\_proj}, \texttt{up\_proj}, \texttt{down\_proj}). For Mixtral-8x7B we compress only \texttt{gate\_proj} and \texttt{up\_proj} and keep \texttt{down\_proj} at full rank; with Mixtral's $d_2/d_1\approx 3.5$ aspect ratio the down-projection ($14336\to4096$) is the most error-sensitive slice of the FFN and we observed materially better downstream scores when it was left uncompressed at matched total storage. The reported total compression ratio accounts for this asymmetry on Mixtral.

\paragraph{Bilinear gauge-fixing.}
After training, each $(\mathbf{U}_{i},\mathbf{V}_{i})$ pair is gauge-fixed (\cref{sec:gauge-fixing}) so the resulting proxy can be saved as a standard HuggingFace MoE checkpoint with no inference-time changes.

\paragraph{Per-model hyperparameters.}
Table~\ref{tab:slbf-hparams} lists the model- and compression-specific settings. The bottleneck rank $k$ is chosen for each compression setting to match the target total-model parameter reduction. We use $\eta=0.03$ and $\rho=0.10$ for Mixtral rather than the default scaling rule, due to its unusually large $d_2/d_1$ ratio. We train for 8,000 iterations on Mixtral and Qwen3.5-122B, whose reconstruction losses had plateaued by that point.
A small number of (layer, projection) pairs exhibit optimization failures. We identify these by manually inspecting the deterministic loss curves, with failures characterized by a large irreversible loss spike or a loss that fails to decrease. For an affected pair, we retrain from the same initialization with $\rho$ halved once (e.g., $0.10\!\to\!0.05$ on Mixtral); no additional retries are performed. The same procedure is applied across all settings. On Mixtral-8x7B at 30\% compression ($k=832$), this affects $6$ out of $32\times2=64$ pairs, while fewer than 1\% of pairs are affected on the other models. A single halving is sufficient in all cases encountered.

\paragraph{MoBE implementation and hyperparameters.}
For MoBE, we directly use the authors' released implementation and recommended hyperparameters. The only modification is on Mixtral-8x7B, where the recommended learning rate did not converge in our setting; we therefore reduce it to $0.03$. All other MoBE settings follow the released implementation without modification.

\section{Ablation Details}

\subsection{Effect of Gauge Fixing on Compression Frontier}
\label{app:gauge_ablation}
 
\begin{figure}
    \centering
    \includegraphics[width=\linewidth]{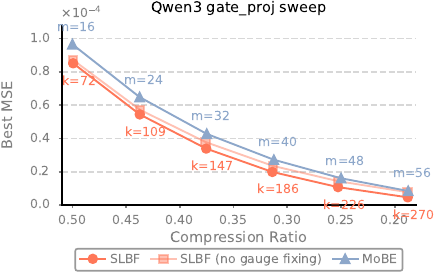}
    \caption{Compression frontier on \texttt{layer 4 gate\_proj} of Qwen3-30B-A3B-Instruct-2507. Best reconstruction MSE versus total compression ratio for SLBF (with gauge fixing), SLBF without gauge fixing, and MoBE. SLBF without gauge fixing already outperforms MoBE across the evaluated sweep; gauge fixing provides an additional gain that widens at higher compression.}
    \label{fig:gauge_ablation}
\end{figure}

We isolate the contribution of bilinear gauge fixing (\cref{sec:gauge-fixing}) by comparing three configurations at varying compression ratios on Qwen3-30B \texttt{gate\_proj}: SLBF with gauge fixing, SLBF without gauge fixing, and MoBE. For each method, we sweep the compression budget by varying the per-basis rank $k$ (SLBF) or the number of bases $m$ (MoBE), training each configuration to convergence. \Cref{fig:gauge_ablation} reports the results. Two conclusions emerge. First, SLBF without gauge fixing already outperforms MoBE across the evaluated compression sweep, showing that the gains are not solely attributable to gauge fixing and providing evidence for the benefit of the low-rank basis design. Second, gauge fixing provides a consistent secondary gain by eliminating $mk^2$ redundant parameters, which are reinvested as higher per-basis rank $k$ at matched budget. This gain widens at more aggressive compression, where the freed parameters constitute a larger fraction of the total budget.

\subsection{Effect of Number of Bases at Matched Budget}
\label{app:m_ablation}

\begin{figure}[h]
\centering
\includegraphics[width=\linewidth]{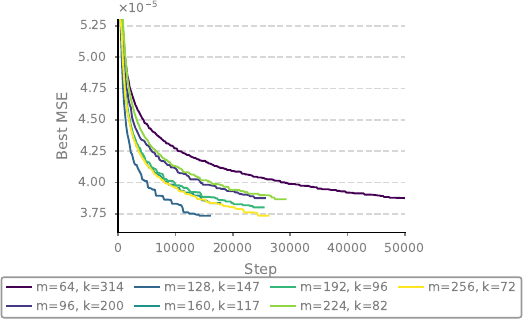}
\caption{Reconstruction loss on \texttt{layer 4 gate\_proj} of Qwen3-30B-A3B-Instruct-2507 at fixed 37.5\% compression, sweeping the number of bases $m$ with per-basis rank $k$ adjusted to match budget. The optimum lies at intermediate $m$; the $m = n = 128$ operating point is near-optimal.}
\label{fig:m_ablation}
\end{figure}

We investigate the tradeoff between the number of bases $m$ and the per-basis rank $k$ under a fixed compression budget. On Qwen3-30B layer 4, \texttt{gate\_proj}, we fix the compression ratio at 37.5\% of the original expert weight tensor and sweep $m \in \{64, 96, 128, 160, 192, 224, 256\}$, adjusting $k$ for each $m$ to match the budget. \Cref{fig:m_ablation} shows the reconstruction loss over training steps. Too few bases ($m = 64$, $k = 314$) converge most slowly and to the highest MSE despite each basis being high-rank; too many bases ($m = 256$, $k = 72$) degrade slightly as each basis becomes too rank-deficient to contribute meaningfully. The optimum lies in the range $m = 160$ to $224$, indicating a genuine tradeoff between basis count and per-basis expressiveness. The operating point $m = n = 128$ used throughout our experiments lies in a strong operating region, and the gains from further increasing $m$ beyond $n$ are modest. We adopt $m = n$, matching the number of bases to experts and simplifying the implementation.

\subsection{Ablation on Downstream Benchmarks}
\label{app:downstream_ablation}

We ablate the effects of the SLBF parameterization, projection scope, and gauge fixing on downstream performance. On Moonlight-16B-A3B at 15\% total compression, comparing MoBE and SLBF under the same \texttt{gate}+\texttt{up} scope, SLBF improves average accuracy from 54.5 to 58.2 and reduces WikiText-2 perplexity from 8.22 to 7.77. Extending SLBF to all three projections at the same total compression ratio further improves the average score to 62.2, while gauge fixing provides an additional gain.

We further conduct an equal-scope control on Qwen3-30B-A3B at 24\% compression. SLBF (\texttt{gate}+\texttt{up}) reaches 83.7 average accuracy compared with 82.8 for MoBE under the same projection scope, while extending SLBF to \texttt{gate}+\texttt{up}+\texttt{down} further improves the score to 84.3. These results separate the two effects: SLBF improves over MoBE even at matched projection scope, while distributing the compression budget across all three projections provides an additional gain.

This finding motivates the projection scope used in \Cref{tab:slbf-hparams}. For Qwen3, Qwen3.5, Moonlight, and Gemma4, we compress all three FFN projections. The original MoBE configuration compresses gate and up projections while leaving down unchanged, following prior observations that down projections can be more sensitive to compression~\citep{geva2021transformer,meng2022locating,liu2025molae}. For Mixtral-8x7B, whose large aspect ratio ($d_2/d_1=3.5$) makes the down projection particularly sensitive in our experiments, we likewise retain \texttt{down\_proj} at full rank and compress only gate and up.

\begin{table}[h]
\centering
\footnotesize
\begin{tabular}{l|c|cc}
\toprule
\textbf{Configuration} & \textbf{Avg.($\uparrow$)}
& \textbf{WT2 ($\downarrow$)} & \textbf{C4 ($\downarrow$)} \\
\midrule
\multicolumn{4}{c}{\textit{Moonlight-16B-A3B, 15\% compression}} \\
\midrule
Original & 64.3 & 6.93 & 12.77 \\
MoLAE (g+u) & 31.5 & 13.02 & 24.33 \\
MoBE (g+u) & 54.4 & 8.22 & 14.67 \\
SLBF (g+u) & 58.2 & 7.77 & 14.08 \\
SLBF (g+u+d, w/o GF) & 62.2 & 7.54 & 13.66 \\
SLBF (g+u+d, w/ GF) & \textbf{62.5} & \textbf{7.53} & \textbf{13.55} \\
\midrule
\multicolumn{4}{c}{\textit{Qwen3-30B-A3B, 24\% compression}} \\
\midrule
Original & 84.9 & & \\
MoBE (g+u) & 82.8 &  &  \\
SLBF (g+u) & 83.7 &  &  \\
SLBF (g+u+d) & \textbf{84.3} &  &  \\
\bottomrule
\end{tabular}
\caption{Downstream ablation of projection scope and gauge fixing.
The Moonlight results additionally report token-level perplexity on
WikiText-2 and C4. The Qwen3 control isolates projection scope at the
same 24\% total-model compression ratio.}
\label{tab:downstream_ablation}
\end{table}

\subsection{Compression Runtime}
\label{app:runtime}
 
Both MoBE and SLBF solve the same weight-space reconstruction objective; the per-iteration cost is comparable since the dominant operation is the forward pass through the parameterization. On Mixtral-8x7B (\texttt{gate\_proj}), MoBE requires a mean of 94.3 minutes per (layer, projection) pair, while SLBF completes in 40.9 minutes using 40\% of the iterations, enabled by its faster convergence (\Cref{fig:slbf-convergence}). Since each (layer, projection) pair is trained independently with no cross-layer coupling, the procedure is embarrassingly parallel.

\subsection{Inference Analysis}
\label{app:inference}
\paragraph{Inference cost reduction under gauge-fixed storage}\label{app:gauge-fixing-inference}

Beyond removing $k^2$ stored parameters per basis (Sec.~\ref{sec:gauge-fixing}), the fixed identity block in $\hat{\mathbf{U}}$ also reduces the matrix-multiplication cost when materializing each basis at inference. Writing the product in block form,
\begin{equation}
\hat{\mathbf{U}}\hat{\mathbf{V}}
= \begin{bmatrix} \mathbf{I}_k \\ \mathbf{U}_b \mathbf{U}_a^{-1} \end{bmatrix} \hat{\mathbf{V}}
= \begin{bmatrix} \hat{\mathbf{V}} \\[2pt] (\mathbf{U}_b \mathbf{U}_a^{-1})\, \hat{\mathbf{V}} \end{bmatrix},
\label{eq:gauge-fixed-mm}
\end{equation}
the top $k$ rows of the basis are simply a copy of $\hat{\mathbf{V}}$, requiring no multiplication, while only the remaining $(r-k)$ rows need an explicit matrix product. The multiply-add cost per basis drops from $r\,k\,d$ to $(r-k)\,k\,d$, a saving of $k^2 d$ operations. Aggregated over $m$ shared bases this is $m\,k^2\,d$ multiply-adds per projection per MoE layer. On Mixtral-8x7B with $m=8$, $k=832$, $d=14336$ the saving is about $20\%$ of the unoptimized basis reconstruction cost per projection per layer.

In practice the $k$ pivot rows in $\hat{\mathbf{U}}$ sit at arbitrary positions selected by column-pivoted QR rather than literally the top of the matrix. We therefore store the pivot indices alongside $\mathbf{U}_b\mathbf{U}_a^{-1}$ and $\hat{\mathbf{V}}$, and assemble each materialized basis by (i) placing $\hat{\mathbf{V}}$ at the pivot rows of the output via an index-scatter, and (ii) computing the remaining $(r-k)$ rows from a smaller batched matrix product on $\mathbf{U}_b\mathbf{U}_a^{-1}$ and $\hat{\mathbf{V}}$.

\paragraph{Runtime storage and throughput}

We evaluate SLBF's runtime-decompressed mode on Mixtral-8x7B against an uncompressed baseline that uses the same per-expert dispatch path\footnote{Both variants share vLLM's stock attention backend~\cite{kwon2023vllm} and use a naive (per-expert loop) MoE implementation. This isolates the cost of basis reconstruction from any optimization gap between fused- and loop-based MoE kernels.}. Both run on 2$\times$ NVIDIA H100 80\,GB (pipeline parallelism PP$=2$, BF16) under vLLM with \texttt{enforce\_eager=True} and \texttt{gpu\_memory\_utilization}$=0.85$ (an effective GPU-memory budget of $\sim$136\,GiB across both ranks).
Table~\ref{tab:runtime} summarizes the trade-off. SLBF reduces on-device model weights by $30\%$; the freed memory becomes KV-cache budget ($+53\%$), so the maximum number of concurrent requests at the configured 4416-token sequence length grows from $75.4$ to $115.5$. At a high-concurrency operating point of 96 prompts (the regime where the uncompressed baseline reaches its KV-cache ceiling), the SLBF throughput is within $12\%$ of the uncompressed baseline (ratio of $0.88\times$), indicating that the runtime-decompression overhead becomes a small fraction of the total compute once it is amortized across enough in-flight tokens.

\begin{table}[h]
\centering
\small
\begin{tabular}{lccc}
\toprule
Metric & Orig & SLBF & Change \\
\midrule
Weights (GiB)               & 87.5  & 61.5  & $-30\%$ \\
KV-cache budget (GiB)             & 40.9  & 62.5  & $+53\%$ \\
KV-cache size (tokens)            & 333k & 510k & $+53\%$ \\
Max concurrent requests   & 75.4  & 115.5 & $+53\%$ \\
Throughput (tok/s)   & 9{,}469 & 8{,}319 & $0.88\times$ \\
\bottomrule
\end{tabular}
\caption{SLBF runtime decompression vs.\ an uncompressed baseline on
Mixtral-8x7B (2$\times$H100 80\,GB, PP$=2$, BF16, vLLM 0.17.1). Memory figures are
summed across both PP ranks. Throughput is reported at the operating point at
which the uncompressed baseline begins to saturate its KV-cache budget.}
\label{tab:runtime}
\end{table}

\section{Source and licenses}
\paragraph{Base-model sources and licenses.}
All base models are obtained from their official HuggingFace repositories and used under their respective public licenses. We release no model weights ourselves; SLBF reconstructions and downstream evaluations operate locally and respect each base model's terms of use. Specifically, we use \href{https://huggingface.co/mistralai/Mixtral-8x7B-v0.1}{Mixtral-8x7B-v0.1} (Apache 2.0), \href{https://huggingface.co/Qwen/Qwen3-30B-A3B-Instruct-2507}{Qwen3-30B-A3B-Instruct-2507} (Apache 2.0), \href{https://huggingface.co/Qwen/Qwen3.5-122B-A10B}{Qwen3.5-122B-A10B} (Apache 2.0), \href{https://huggingface.co/moonshotai/Moonlight-16B-A3B-Instruct}{Moonlight-16B-A3B-Instruct} (MIT), and \href{{https://huggingface.co/google/gemma-4-26b-a4b-it}}{Gemma-4-26B-A4B-it} (\href{https://ai.google.dev/gemma/docs/gemma_4_license}{Gemma license}).

\paragraph{Dataset.}
For baseline methods (REAP, HC-SMoE, MC-SMoE, Sub-MoE, etc.), calibration uses 1024 sequences of 2048 tokens from the \href{https://huggingface.co/datasets/allenai/c4}{C4 corpus}~\citep{raffel2020c4} (ODC-By~1.0). All downstream evaluations are run via the EleutherAI \href{https://github.com/EleutherAI/lm-evaluation-harness}{\texttt{lm-evaluation-harness}}~\citep{eval-harness}, which we use under Apache~2.0 along with each task's documented license. Code and evaluation scripts are available in our \href{https://github.com/ctxGou/MoE-Compression}{public repository}.

\subsection{Evaluation Details}
\label{app:evaluation}

\paragraph{Implementation.}
All evaluations use the vLLM backend with \texttt{lm-evaluation-harness} v0.4.11.
Qwen3, Moonlight, and Mixtral use vLLM v0.17.1; Gemma-4 and Qwen3.5 use v0.19.1.
Original and compressed checkpoints use identical settings.

\paragraph{Tasks and prompting.}
For the instruction-tuned models, we evaluate \texttt{mmlu\_flan\_cot\_zeroshot},
\texttt{gpqa\_diamond\_cot\_zeroshot}, C-Eval, CMMLU, \texttt{gsm8k},
\texttt{arc\_challenge\_chat}, 
\texttt{leaderboard\_ifeval}, \texttt{mbpp}, and \texttt{HumanEval}.
C-Eval and CMMLU use \texttt{ceval-valid} and \texttt{cmmlu} for
Qwen3, Moonlight, and Qwen3.5, and \texttt{ceval-valid-chat} and
\texttt{cmmlu-chat} for Gemma-4.
All tasks are zero-shot except \texttt{gsm8k} (5-shot) and
\texttt{mbpp} (3-shot).
Generative tasks use each model's tokenizer-provided chat template;
few-shot examples are separate user/assistant turns, while
log-likelihood tasks disable chat formatting.

\paragraph{Decoding settings.}
Qwen3 and Moonlight use greedy decoding.
Gemma-4 uses temperature $1.0$, top-$p$ $0.95$, and top-$k$ $64$;
Qwen3.5 uses temperature $1.0$, top-$p$ $0.95$, top-$k$ $20$,
min-$p$ $0$, and presence penalty $1.5$.
Maximum output lengths are 4,096 tokens for Qwen3, task-dependent
1,024--6,144 tokens for Moonlight, task-dependent 32--32,768 tokens
for Gemma-4, and 32,768 tokens for Qwen3.5.

\end{document}